\documentclass{article}
\pdfoutput=1
    \PassOptionsToPackage{numbers, compress}{natbib}

    \usepackage[preprint]{neurips_2020}

\usepackage[utf8]{inputenc} 
\usepackage[T1]{fontenc}    
\usepackage{hyperref}       
\usepackage{url}            
\usepackage{booktabs}       
\usepackage{amsfonts}       
\usepackage{amsmath}
\usepackage{nicefrac}       
\usepackage{microtype}      
\usepackage{graphicx}       
\usepackage{xcolor}         
\usepackage{todonotes}
\usepackage{multirow}       
\graphicspath{{Images/}}    

\title{Detecting Hate Speech in Memes Using Multimodal Deep Learning Approaches: Prize-winning solution to Hateful Memes Challenge}

\author{%
  Riza Velioglu \\
  Technical Faculty\\
  Bielefeld University\\
  Universitätsstraße 25, 33615 Bielefeld, Germany \\
  \texttt{rvelioglu@techfak.uni-bielefeld.de} \\
  \And
  Jewgeni Rose \\
  Smart Data Analytics \\
  Computer Science III, University of Bonn, Germany \\
  \texttt{jewgeni.rose@gmail.com} \\
}

\begin{document}
\maketitle

\begin{abstract}
  Memes on the Internet are often harmless and sometimes amusing. However, by using certain types of images, text, or combinations of both, the seemingly harmless meme becomes a multimodal type of hate speech -- a \textit{hateful meme}. The Hateful Memes Challenge\footnote{\url{https://www.drivendata.org/competitions/70/hateful-memes-phase-2/}} is a first-of-its-kind competition which focuses on detecting hate speech in multimodal memes and it proposes a new data set containing 10,000+ new examples of multimodal content. We utilize VisualBERT -- which meant to be the ``BERT of vision and language'' -- that was trained multimodally on images and captions and apply Ensemble Learning. Our approach achieves 0.811 AUROC with an accuracy of 0.765 on the challenge test set and placed third out of 3,173 participants in the Hateful Memes Challenge\footnote{HateDetectron at \url{https://www.drivendata.org/competitions/70/hateful-memes-phase-2/leaderboard/}}. The code is available at 
    \url{https://github.com/rizavelioglu/hateful_memes-hate_detectron}
\end{abstract}

\section{Introduction}
\label{intro}

  Memes have gained huge popularity over the past years, resulting in over 180m posts on different social media platforms until 2018~\cite{zannettou2018origins}. Although memes are oftentimes harmless and generated especially for humorous purposes, they have also been used to produce and disseminate hate speech in toxic communities. Hate Speech (HS) is a direct attack on people based on race, ethnicity, national origin, religious affiliation, sexual orientation, sex, gender, and serious disease or disability~\cite{fb} -- a growing problem in modern society. Giant tech companies, such as Facebook, own platforms where millions of users log in daily and they are obliged to remove a tremendous amount of HS to protect their users. According to Mike Schroepfer, Facebook CTO, they took an action on \emph{9.6 million} pieces of content for violating their HS policies in the first quarter of 2020~\cite{fbCTO}. This amount of malicious content cannot be tackled by having humans inspect every sample. Consequently, machine learning and in particular deep learning techniques are required to alleviate the extensiveness of online hate speech. Detecting hate speech in memes is challenging due to the multimodal nature of memes (usually image+text). Therefore, these techniques have to process the content the way humans do: holistically. When viewing a meme, a human would not think about the words and the picture independently; but understand the \textit{combined} meaning. Moreover, while the visual and linguistic information of a meme is typically neutral or funny individually, their combination may result in a hateful meme.
  
  A recent study shows that state-of-the-art methods for hate speech detection in multimodal memes perform poorly compared to humans: \emph{64.73\%} vs. \emph{84.7\%} accuracy~\cite{kiela2020hateful}. To catalyze sophisticated research in this area, Facebook AI launched the Hateful Memes Challenge and published a dataset containing more than 10,000 newly created multimodal memes~\cite{kiela2020hateful}. Multimodal tasks reflect many real-world problems, including how humans perceive and understand the world around them. 
  
  There has been a surge of interest in multimodal problems since 2015 in visual question answering~\cite{antol2015vqa, gurari2018vizwiz}, image captioning~\cite{krishna2017visual, chen2015microsoft}, speech recognition~\cite{srinivasan2020multimodal, paraskevopoulos2020multiresolution} and beyond. But it is not always clear to what extent genuinely multimodal reasoning and understanding are needed to solve current challenges. For instance, for some datasets language can unintentionally impose strong priors, which might result in a remarkable performance, without any understanding of the visual content. The Hateful Memes challenge design and dataset are created to encourage and measure truly multimodal understanding and reasoning of the models. A key point to achieve this are the so-called ``benign confounders'' (also called \textit{contrastive}~\cite{gardner2020evaluating} or \textit{counterfactual}~\cite{kaushik2019learning} examples) which addresses the risk of exploiting unimodal priors by models: for every hateful meme, there are alternative images or text that flip the label to not-hateful. Such image and text confounders require multimodal reasoning to classify the original meme and its confounders correctly. Thus, making the dataset challenging and appropriate for testing the true multimodality of a model.

  In the following, we analyze the challenge dataset and describe our prize-winning solution that placed third among 3,173 participants in the Hateful Memes Challenge in detail. Our solution achieves \emph{0.811} AUROC with an accuracy of \emph{0.765} on the challenge test set, which improves all the benchmark models~\cite{kiela2020hateful}, including the state-of-the-art models at that time, such as ViLBERT~\cite{lu2019vilbert} (trained on Conceptual Captions~\cite{sharma2018conceptual}) and VisualBERT~\cite{li2019visualbert} (trained on COCO~\cite{chen2015microsoft}). Nevertheless, the accuracy is still behind humans with a mentionable gap, highlighting the need for progress in multimodal research.

\section{Problem Statement}
\label{prob}

  The Hateful Memes dataset is not created for training models from scratch, but to fine-tune and test large-scale, pre-trained multimodal models. Thus, the size of the dataset (10K images) is small compared to datasets such as Visual Genome (108K)~\cite{krishna2017visual}, COCO (330K)~\cite{chen2015microsoft}, and Conceptual Captions (3.3M)~\cite{sharma2018conceptual}. The dataset is split into three sets: a train set of 8.500 samples, a dev set of 500 samples, and a test set of 1.000 samples. In addition to this ``seen'' test set, a new test set consisting of 2.000 samples has been published where the winners are determined according to their performance on this ``unseen'' test set. The area under the receiver operating characteristic curve (ROC AUC)~\cite{bradley1997use} has been selected as the measure of performance, which is given by the following formula:
  \begin{equation}
        AUROC = \int_{x=0}^{1} \texttt{TPR}(\texttt{FPR}^{-1}(x))dx
  \end{equation}
  The labels indicating whether a meme is hateful or not-hateful are provided within the dataset, hence the task can be cast as a binary classification problem.

\section{Methods}
\label{methods}

\begin{figure} 
  \includegraphics[width=\textwidth]{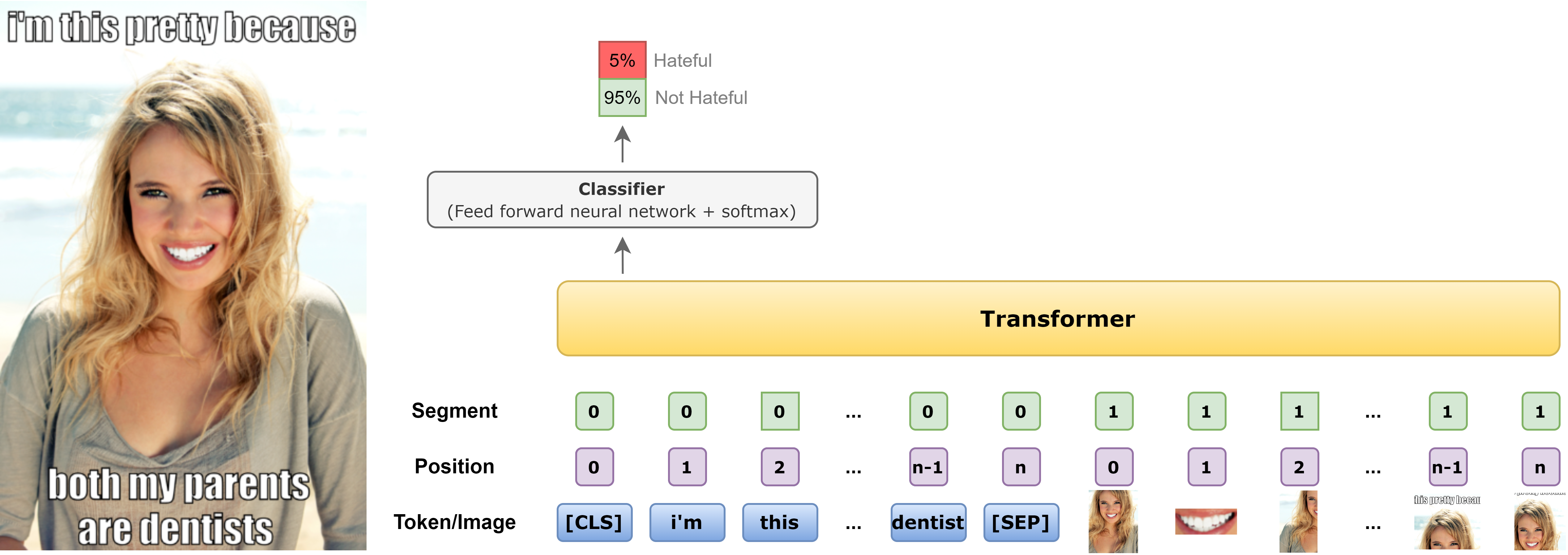}
  \caption{An example meme sampled from the dataset (left), and an illustration of the multimodal transformer architecture (right). Image regions and language are combined with a Transformer to allow the self-attention to discover implicit alignments between language and vision.}
  \centering
  \label{fig:model}
\end{figure}

The solution comprises VisualBERT~\cite{li2019visualbert}, a multimodal BERT for vision-and-language approach. Figure~\ref{fig:model} illustrates an overview of the architecture. The approach can be divided into four sections: dataset expansion, image encoding, training, and ensemble learning. Next, we will provide details of our solution.

\subsection{Dataset Expansion}
  More data delivers stable learning and brings better scores. Thus, we searched for additional data sources in order to grow the dataset size and as a result, we expanded the training data by 428 additional memes.
  
  \paragraph{Unused data in dev seen}
  We found that there are 500 samples in the seen dev set and 540 samples in the unseen dev set. By comparing the memes by their IDs, we identified 400 overlapping samples: $|\{dev\_seen\} \cap \{dev\_seen\}|=400$, which means that there are 100 samples that are not in dev unseen: $|\{dev\_seen\} \backslash \{dev\_seen\}|=100$. We added these 100 samples to the training data and evaluated the trained model on dev unseen. 

  \paragraph{Memotion Dataset}
  The Memotion Dataset~\cite{sharma2020semeval} is an open-sourced dataset containing 14K annotated memes with human-annotated labels, namely \emph{sentiment}(positive, negative, neutral), type of \emph{emotion}(sarcastic, funny, offensive, motivational). For instance, a meme could be annotated as \emph{Not Funny, Very Twisted, Hateful Offensive, Not Motivational}. After an exploratory analysis of the dataset, we argue that the majority of the samples are wrongly labeled. Therefore, we manually re-labeled a part of the dataset. We picked memes that are similar to the ones in the Hateful Memes dataset considering the meme style and design of the challenge dataset. After cherry-picking the ``similar’’ memes, we added 328 new memes to the training data.

\subsection{Image Encoding}
\begin{figure} 
  \centering
  \includegraphics[width=0.4\textwidth]{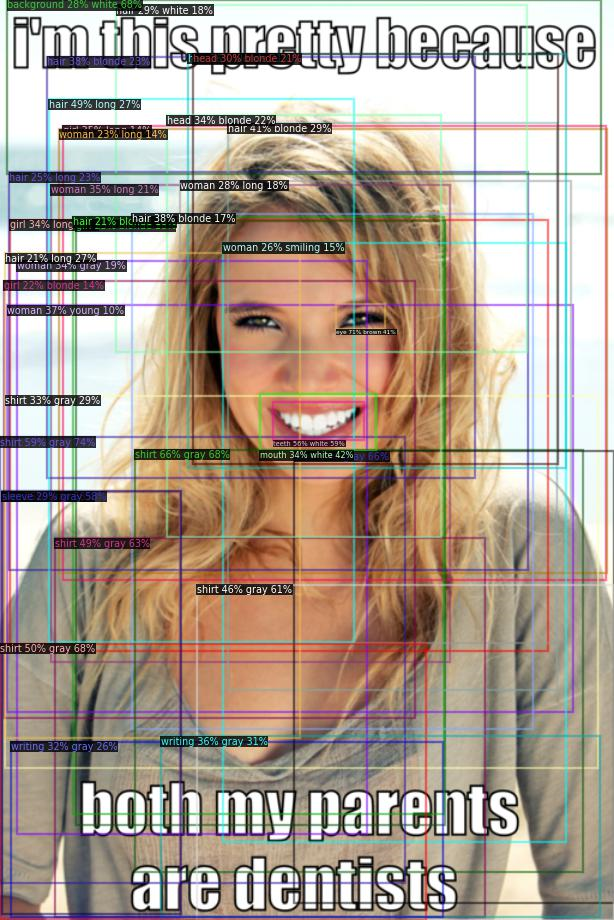}
  \caption{An example of a processed image where the boxes are extracted by Mask-RCNN. Originally 100 boxes are extracted per image but for plotting purposes only 36 boxes are shown.}
  \label{fig:processed}
\end{figure}

  For every image we extract 100 boxes of 2048$\mathcal{D}$ region-based image features from a \texttt{fc6} layer of a ResNeXT-152 based Mask-RCNN model~\cite{he2017mask}, trained on Visual Genome~\cite{krishna2017visual} with the attribute prediction loss following~\cite{anderson2018bottom}. Figure~\ref{fig:processed} shows an example of a processed image. We project the visual embeddings into the textual embedding space before passing them through the transformer layers. We learn weights $W_n \in \mathbb{R}^{PxD}$ to project each of the 100 image embeddings to $D$-dimensional token input embedding space:
    \begin{equation}
        I_n = W_n f(img,n),
  \end{equation}
  where $P=2048$, $D=768$, and $f(\cdot,n)$ is the output of the $n$-th fully-connected layer in the image encoder.

\subsection{Training}
  \paragraph{Pre-training}
      VisualBERT is originally pre-trained on COCO image caption dataset~\cite{chen2015microsoft}, but in our experiments we noticed that the model pre-trained on Conceptual Captions~\cite{sharma2018conceptual} achieves noticeably better scores. Therefore, we conducted our research on the latter model which is provided by MMF: a framework for vision-and-language multimodal research from Facebook AI Research (FAIR)~\cite{singh2020mmf}.

  \paragraph{Fine-tuning}
      We fine-tune the pre-trained VisualBERT model on the aggregated training set and evaluate it on dev unseen set. 

  \paragraph{Classification}
      We use the first output of the final layer as the input to a classification layer $\texttt{clf}(x) = Wx+b$ where $W \in \mathbb{R}^{DxC}$, with $D$ as the transformer dimensionality and $C$ as the number of classes (also see Figure~\ref{fig:model}). We apply a softmax on the logits and train with binary cross-entropy loss.

\subsection{Ensemble Learning}
  The idea of ensemble learning is to combine the predictions of multiple base models in order to improve generalizability and robustness over a single model. Specifically, we use Majority Voting technique (also known as Hard Voting or Voting Classifier) which combines different classifiers and use a majority vote to predict the class labels. The resulting classifier is oftentimes useful for a variety of equally well performing model as to balance out their individual weaknesses. Consequently, it achieves better performance than any single model used in the ensemble.

  We constructed a hyper-parameter search that resulted in multiple models having different AUROC scores on dev unseen set. After sorting them by the AUROC score, the top 27 models are selected for ensemble learning as shown in Table~\ref{table:ensemble} (the number of models is chosen arbitrarily). Then, predictions are collected from each of the models and the majority voting technique is applied: the class of a data point is determined by the majority voted class. Besides, in order to calculate AUROC, the probability that a data point is assigned to a class has to be determined: If the majority voted class is 1 (hateful), then the probability is the maximum among all the 27 models and minimum if it is class 0 (not hateful).

\section{Experiments and Results}
\label{experiments}
\begin{table}
  \caption{Ensemble models performances derived from VisualBERT CC}
  \label{table:ensemble}
  \centering
  \begin{tabular}{c|cc}
  \toprule
\multirow{2}{*}{\textbf{ID}}    & \multicolumn{2}{c}{\textbf{Validation}} \\
                                & Acc.             & AUROC  \\
\midrule
1                               & 70.93            & \textbf{75.21 } \\                
2                               & 69.63            & 75.16         \\                  
3                               & 70.74            & 75.02         \\                  
$\cdots$                             & $\cdots$              & $\cdots$           \\
25                              & 70.56            & 73.76         \\
26                              & 70.93            & 73.75         \\
27                              & 69.81            & 73.68         \\
\bottomrule
\end{tabular}
\end{table}
\begin{table}
  \caption{Model performance}
  \label{table:model}
  \centering
  \begin{tabular}{ll|cc|cc}
    \toprule
                                               &                 & \multicolumn{2}{c|}{\textbf{Validation}} & \multicolumn{2}{c}{\textbf{Test}}  \\
\textbf{Type}                                  & \textbf{Model}  & Acc.                & AUROC              & Acc.             & AUROC           \\
    \midrule
                                               & Human           & -                   & -                  & 84.70            & 82.65           \\
    \midrule
\multicolumn{1}{c}{\multirow{4}{*}{Baselines}} & ViLBERT         & 62.20               & 71.13              & 62.30            & 70.45           \\
\multicolumn{1}{c}{}                           & VisualBERT      & 62.10               & 70.60              & 63.20            & 71.33           \\
\multicolumn{1}{c}{}                           & ViLBERT CC      & 61.40               & 70.07              & 61.10            & 70.03           \\
\multicolumn{1}{c}{}                           & VisualBERT COCO & 65.06               & 73.97              & 64.73            & 71.41           \\
    \midrule
\multicolumn{1}{c}{\multirow{2}{*}{Ours}}      & Ensemble        & -                   & -                  & \textbf{76.50}   & \textbf{81.08}  \\
\multicolumn{1}{c}{}                           & Best ensemble model& \textbf{70.93}   & \textbf{75.21}     & -                & -               \\
    \bottomrule
  \end{tabular}
\end{table}

  Majority Voting boosted both AUROC and accuracy by \textbf{2.5\%}. We argue that this technique successfully applies ensemble learning and generates one strong model from multiple ‘weak’ models -- in analogy to the idea of 'bringing the experts of the experts together'. Imagine that one model is very good at -- in other words, an expert -- detecting hate speech towards women, but might not be an expert in detecting hate speech towards religion. Then, we might have another expert whose expertise is just the opposite. By using the majority voting technique, we bring such experts all together and benefit from them as a whole. The results are shown in Table~\ref{table:model}.

\section{Conclusion}
\label{discussion}
  We proposed an approach detecting hate speech in internet memes multimodally, i.e. considering visual and textual information holistically. We took part in the Hateful Memes Challenge and placed third out of 3,173 participants. Our approach utilizes a pre-trained VisualBERT (a BERT of vision and language), fine-tuned on an expanded train dataset, finally applying Majority Voting over the 27 best models. Our approach achieves 0.811 AUROC with an accuracy of 0.765 on the challenge test set, which is a considerable result but also shows that we are still far from the accuracy of human judgement.   

\small
\bibliographystyle{unsrtnat}
\bibliography{sample}

\begin{thebibliography}{20}
\providecommand{\natexlab}[1]{#1}
\providecommand{\url}[1]{\texttt{#1}}
\expandafter\ifx\csname urlstyle\endcsname\relax
  \providecommand{\doi}[1]{doi: #1}\else
  \providecommand{\doi}{doi: \begingroup \urlstyle{rm}\Url}\fi

\bibitem[Zannettou et~al.(2018)Zannettou, Caulfield, Blackburn, De~Cristofaro,
  Sirivianos, Stringhini, and Suarez-Tangil]{zannettou2018origins}
Savvas Zannettou, Tristan Caulfield, Jeremy Blackburn, Emiliano De~Cristofaro,
  Michael Sirivianos, Gianluca Stringhini, and Guillermo Suarez-Tangil.
\newblock On the origins of memes by means of fringe web communities.
\newblock In \emph{Proceedings of the Internet Measurement Conference 2018},
  pages 188--202, 2018.

\bibitem[{Facebook}(2020)]{fb}
{Facebook}.
\newblock {Community Standards}.
\newblock \url{https://www.facebook.com/communitystandards/hate_speech}, 2020.

\bibitem[S.Perry(2020)]{fbCTO}
Tekla S.Perry.
\newblock Q\&a: Facebook’s cto is at war with bad content, and ai is his best
  weapon.
\newblock
  \url{https://spectrum.ieee.org/computing/software/qa-facebooks-cto-is-at-war-with-bad-content-and-ai-is-his-best-weapon},
  7 2020.

\bibitem[Kiela et~al.(2020)Kiela, Firooz, Mohan, Goswami, Singh, Ringshia, and
  Testuggine]{kiela2020hateful}
Douwe Kiela, Hamed Firooz, Aravind Mohan, Vedanuj Goswami, Amanpreet Singh,
  Pratik Ringshia, and Davide Testuggine.
\newblock The hateful memes challenge: Detecting hate speech in multimodal
  memes.
\newblock \emph{arXiv preprint arXiv:2005.04790}, 2020.

\bibitem[Antol et~al.(2015)Antol, Agrawal, Lu, Mitchell, Batra,
  Lawrence~Zitnick, and Parikh]{antol2015vqa}
Stanislaw Antol, Aishwarya Agrawal, Jiasen Lu, Margaret Mitchell, Dhruv Batra,
  C~Lawrence~Zitnick, and Devi Parikh.
\newblock Vqa: Visual question answering.
\newblock In \emph{Proceedings of the IEEE international conference on computer
  vision}, pages 2425--2433, 2015.

\bibitem[Gurari et~al.(2018)Gurari, Li, Stangl, Guo, Lin, Grauman, Luo, and
  Bigham]{gurari2018vizwiz}
Danna Gurari, Qing Li, Abigale~J Stangl, Anhong Guo, Chi Lin, Kristen Grauman,
  Jiebo Luo, and Jeffrey~P Bigham.
\newblock Vizwiz grand challenge: Answering visual questions from blind people.
\newblock In \emph{Proceedings of the IEEE Conference on Computer Vision and
  Pattern Recognition}, pages 3608--3617, 2018.

\bibitem[Krishna et~al.(2017)Krishna, Zhu, Groth, Johnson, Hata, Kravitz, Chen,
  Kalantidis, Li, Shamma, et~al.]{krishna2017visual}
Ranjay Krishna, Yuke Zhu, Oliver Groth, Justin Johnson, Kenji Hata, Joshua
  Kravitz, Stephanie Chen, Yannis Kalantidis, Li-Jia Li, David~A Shamma, et~al.
\newblock Visual genome: Connecting language and vision using crowdsourced
  dense image annotations.
\newblock \emph{International journal of computer vision}, 123\penalty0
  (1):\penalty0 32--73, 2017.

\bibitem[Chen et~al.(2015)Chen, Fang, Lin, Vedantam, Gupta, Doll{\'a}r, and
  Zitnick]{chen2015microsoft}
Xinlei Chen, Hao Fang, Tsung-Yi Lin, Ramakrishna Vedantam, Saurabh Gupta, Piotr
  Doll{\'a}r, and C~Lawrence Zitnick.
\newblock Microsoft coco captions: Data collection and evaluation server.
\newblock \emph{arXiv preprint arXiv:1504.00325}, 2015.

\bibitem[Srinivasan et~al.(2020)Srinivasan, Sanabria, Metze, and
  Elliott]{srinivasan2020multimodal}
Tejas Srinivasan, Ramon Sanabria, Florian Metze, and Desmond Elliott.
\newblock Multimodal speech recognition with unstructured audio masking.
\newblock \emph{arXiv preprint arXiv:2010.08642}, 2020.

\bibitem[Paraskevopoulos et~al.(2020)Paraskevopoulos, Parthasarathy, Khare, and
  Sundaram]{paraskevopoulos2020multiresolution}
Georgios Paraskevopoulos, Srinivas Parthasarathy, Aparna Khare, and Shiva
  Sundaram.
\newblock Multiresolution and multimodal speech recognition with transformers.
\newblock \emph{arXiv preprint arXiv:2004.14840}, 2020.

\bibitem[Gardner et~al.(2020)Gardner, Artzi, Basmova, Berant, Bogin, Chen,
  Dasigi, Dua, Elazar, Gottumukkala, et~al.]{gardner2020evaluating}
Matt Gardner, Yoav Artzi, Victoria Basmova, Jonathan Berant, Ben Bogin, Sihao
  Chen, Pradeep Dasigi, Dheeru Dua, Yanai Elazar, Ananth Gottumukkala, et~al.
\newblock Evaluating nlp models via contrast sets.
\newblock \emph{arXiv preprint arXiv:2004.02709}, 2020.

\bibitem[Kaushik et~al.(2019)Kaushik, Hovy, and Lipton]{kaushik2019learning}
Divyansh Kaushik, Eduard Hovy, and Zachary~C Lipton.
\newblock Learning the difference that makes a difference with
  counterfactually-augmented data.
\newblock \emph{arXiv preprint arXiv:1909.12434}, 2019.

\bibitem[Lu et~al.(2019)Lu, Batra, Parikh, and Lee]{lu2019vilbert}
Jiasen Lu, Dhruv Batra, Devi Parikh, and Stefan Lee.
\newblock Vilbert: Pretraining task-agnostic visiolinguistic representations
  for vision-and-language tasks.
\newblock In \emph{Advances in Neural Information Processing Systems}, pages
  13--23, 2019.

\bibitem[Sharma et~al.(2018)Sharma, Ding, Goodman, and
  Soricut]{sharma2018conceptual}
Piyush Sharma, Nan Ding, Sebastian Goodman, and Radu Soricut.
\newblock Conceptual captions: A cleaned, hypernymed, image alt-text dataset
  for automatic image captioning.
\newblock In \emph{Proceedings of the 56th Annual Meeting of the Association
  for Computational Linguistics (Volume 1: Long Papers)}, pages 2556--2565,
  2018.

\bibitem[Li et~al.(2019)Li, Yatskar, Yin, Hsieh, and Chang]{li2019visualbert}
Liunian~Harold Li, Mark Yatskar, Da~Yin, Cho-Jui Hsieh, and Kai-Wei Chang.
\newblock Visualbert: A simple and performant baseline for vision and language.
\newblock \emph{arXiv preprint arXiv:1908.03557}, 2019.

\bibitem[Bradley(1997)]{bradley1997use}
Andrew~P Bradley.
\newblock The use of the area under the roc curve in the evaluation of machine
  learning algorithms.
\newblock \emph{Pattern recognition}, 30\penalty0 (7):\penalty0 1145--1159,
  1997.

\bibitem[Sharma et~al.(2020)Sharma, Bhageria, Scott, PYKL, Das, Chakraborty,
  Pulabaigari, and Gamback]{sharma2020semeval}
Chhavi Sharma, Deepesh Bhageria, William Scott, Srinivas PYKL, Amitava Das,
  Tanmoy Chakraborty, Viswanath Pulabaigari, and Bjorn Gamback.
\newblock Semeval-2020 task 8: Memotion analysis--the visuo-lingual metaphor!
\newblock \emph{arXiv preprint arXiv:2008.03781}, 2020.

\bibitem[He et~al.(2017)He, Gkioxari, Doll{\'a}r, and Girshick]{he2017mask}
Kaiming He, Georgia Gkioxari, Piotr Doll{\'a}r, and Ross Girshick.
\newblock Mask r-cnn.
\newblock In \emph{Proceedings of the IEEE international conference on computer
  vision}, pages 2961--2969, 2017.

\bibitem[Anderson et~al.(2018)Anderson, He, Buehler, Teney, Johnson, Gould, and
  Zhang]{anderson2018bottom}
Peter Anderson, Xiaodong He, Chris Buehler, Damien Teney, Mark Johnson, Stephen
  Gould, and Lei Zhang.
\newblock Bottom-up and top-down attention for image captioning and visual
  question answering.
\newblock In \emph{Proceedings of the IEEE conference on computer vision and
  pattern recognition}, pages 6077--6086, 2018.

\bibitem[Singh et~al.(2020)Singh, Goswami, Natarajan, Jiang, Chen, Shah,
  Rohrbach, Batra, and Parikh]{singh2020mmf}
Amanpreet Singh, Vedanuj Goswami, Vivek Natarajan, Yu~Jiang, Xinlei Chen, Meet
  Shah, Marcus Rohrbach, Dhruv Batra, and Devi Parikh.
\newblock Mmf: A multimodal framework for vision and language research, 2020.

\end{thebibliography}

\end{document}